\documentclass[preprint,12pt]{elsarticle}
\usepackage{ulem}

\usepackage{amssymb}
\usepackage{amsmath}
\usepackage{amsmath}
\usepackage{float}
\usepackage{algorithm}
\usepackage{algorithmic}
\usepackage{booktabs} 
\usepackage{graphicx}
\usepackage{verbatim}
\usepackage{subcaption}
\usepackage{makecell}
\usepackage{xcolor}
\usepackage{multirow}

\journal{Computational Statistics \& Data Analysis}
\newcommand{\praf}[1]{{\footnote{{/* #1 [Firmino] */}}}}
\newcommand{\prafC}[1]{\iffalse{/* #1 [Firmino] */}\fi}
\begin{document}
\begin{frontmatter}



\title{Forecasting Oncology Demand Trends with Boosting-Based Bayesian Conjugate Models}


\author{Ademir Batista dos Santos Neto}
\ead{ademir.neto@ufca.edu.br} 
\affiliation{organization={Universidade Federal do Cariri (UFCA)},
            addressline={Rua Tenente Raimundo Rocha, 1639},
            city={Juazeiro do Norte},
            postcode={63048-080},
            state={Ceará},
            country={Brasil}} 

\author{Paulo Renato Alves Firmino}
\ead{paulo.firmino@ufca.edu.br} 
\affiliation{organization={Universidade Federal do Cariri (UFCA)},
            addressline={Rua Tenente Raimundo Rocha, 1639},
            city={Juazeiro do Norte},
            postcode={63048-080},
            state={Ceará},
            country={Brasil}}

\author{Tiago A. E. Ferreira}
\ead{tiago.espinola@ufrpe.br} 
\affiliation{organization={Universidade Federal Rural de Pernambuco (UFRPE)},
            addressline={Rua Dom Manoel de Medeiros, s/n},
            city={Recife},
            postcode={52171-900},
            state={Pernambuco},
            country={Brasil}}
\begin{abstract}
\prafC{falta uma vírgula antes dos emails}

Accurate trend forecasting in healthcare time series is essential for planning and resource allocation. 
This paper proposes a Bayesian framework for predicting oncology demand trends, modeling weekly appointments as a Poisson process with a Gamma prior to the demand rate. 
To enhance adaptability and capture persistent directional patterns, we incorporate a residual-based boosting mechanism grounded in a Gamma–Log-Normal conjugate structure. 
This boosting approach allows the model to track both short- and long-term trend shifts while maintaining the analytical tractability of conjugate Bayesian updating. 
The methodology was evaluated on real oncology service data from Cariri, Ceará, Brazil, and compared against established baselines, including linear regression, ARIMA, naïve forecasting, LSTM neural networks, and XGBoost. 
Results showed that the proposed model outperforms competing methods in trend detection accuracy, with gains in terms of percentage of correct direction 38,25\%\prafC{trazer valor da maior razão de POCID do nosso método e do segundo melhor entre as séries .} in relation to the second best approach in some cases.
\end{abstract}

\begin{graphicalabstract}
\end{graphicalabstract}

\begin{highlights}
\item Bayesian framework for boosting Gamma--Poisson models via Gamma--Log--Normal multiplicative errors.
\item Explicit emphasis on \textbf{trend direction} when forecasting oncology demands by using POCID.
\item Closed-form and sequential updating, enabling fast adaptation to nonstationary demand series.
\item Statistically grounded comparison via  Diebold--Mariano test on POCID elements.
\item Real-world evaluation on oncology services (Cariri, Ceará, Brazil), showing significant gains over ARIMA, Linear Regression, Naive, Long Short-Term Memory Artificial Neural Networks, and XGBoost models.
\end{highlights}


\end{frontmatter}



\section{Introduction}
Forecasting healthcare demand trends is critical for planning and optimizing health service provision, particularly in high-impact areas such as oncology \cite{soyiri2013overview}. 
Oncology units frequently face substantial variability in patient volumes driven by delayed diagnoses, individualized treatment protocols, and irregular follow-up visits, which collectively complicate the process of accurately identifying and predicting demand trends, as well as optimizing resource allocation \cite{smittenaar2016cancer}.
Traditional time series models such as Autoregressive Integrated Moving Average (ARIMA) \cite{box2015time} or Exponential Smoothing (ETS) \cite{hyndman2002state} often struggle to capture structural shifts, trend dynamics, or to incorporate prior knowledge about the process \cite{west2006bayesian}. 
In this context, Bayesian models offer a principled framework for online learning and uncertainty quantification, which are particularly advantageous in dynamic healthcare environments, especially when addressing nonstationary patterns and trend prediction \cite{gelman2013bayesian,zhou2022bayesian}.

Gama-Poisson conjugation is widely used in Bayesian forecasting exercises, especially in the healthcare field \cite{minois2017deal, turchetta2024time, sutradhar2008forecasting, dzupire2018poisson, diao2023generalized}.
In turn, boosting is the well known technique of modeling residuals to enhance the prediction \cite{friedman2001greedy}. 
This study proposes a Bayesian inference framework for forecasting weekly oncology service demand, based on the conjugation of Gamma and Poisson distributions to model weekly demand,  together with a Gamma–Log-Normal conjugation applied to the residuals of the base predictions. 
The base model is, therefore, extended with a residual correction mechanism that explicitly accounts for systematic deviations between observed and predicted values. 
In this context, Bayesian models provide a principled framework for sequential learning and uncertainty quantification, 
which is particularly advantageous in dynamic healthcare environments, especially when dealing with nonstationary patterns, trend prediction, or bounded response variables \cite{gelman2013bayesian, zhou2022bayesianbeta}\prafC{fiquei com a impressão de já ter lido esse texto a pouco tempo atra´s ... Evite repetições...}. 
This boosting approach enables the model to adapt dynamically to changes in the data-generating process 
while preserving the analytical tractability of conjugate Bayesian updating.

The rest of the paper is structured as follows. 
Section 2 describes the theoretical background of conjugation models, especially about Gamma-Poisson and Gamma-Log-Normal. 
Section 3 presents the proposed methodology, with a focus on forecasting with residual adjustment. 
Section 4 details the experimental setup and evaluation measures. 
In Section 5, we report and analyze the results obtained from real oncology demand data, comparing our model with baseline approaches such as linear regression, Naïve forecasting, XGBoost, ARIMA, and Long Short-Term Memory neural networks. 
Finally, Section 6 summarizes the conclusions of the paper.

\section{Conjugate Distributions}

In Bayesian statistics, the process of updating prior beliefs using observed data often leads to challenging posterior distributions that may not admit closed form expressions. 
In these cases, simulation-based methods such as Markov Chain Monte Carlo (MCMC) are commonly used \cite{gelman2013bayesian}.
However, when analytical tractability is desirable, especially in real-time or resource-constrained settings, conjugate priors offer a powerful alternative \cite{murphy2012machine}.

Conjugate priors can be defined as prior distributions that, when combined with a likelihood function from a specific family, yield a posterior distribution in the same family \cite{gelman2013bayesian}.
This property simplifies Bayesian updating and avoids costly numerical integration. 
For instance, if the likelihood is from the exponential family, it is often possible to find a conjugate prior \cite{gutierrez1997exponential}. 
Formally, let the likelihood of the data $x$ conditioned on the parameters $\lambda$ be $p(x \mid \lambda)$ and let the prior be $p(\lambda)$. 
The posterior is then:

\begin{equation}
    p(\lambda \mid x) \propto p(x \mid \lambda) \cdot p(\lambda).
\end{equation}

If the resulting posterior $p(\lambda \mid x)$ belongs to the same family as  $p(\lambda)$, we say that $p(\lambda)$ is a conjugate prior for the likelihood $p(x \mid \lambda)$\cite{degroot2005optimal}\prafC{referencie}.
\prafC{talvez seja melhor usar termos/simbolos similares aos que já adotamos nas subseções a seguir ...}

In this paper, we focus on two classical conjugate pairs for modeling count and predictor residuals time series: respectively \textit{Gamma--Poisson} and \textit{Gamma--Log-Normal}. 
 In this way, let $\mathbf{x}_n = (x_1, \dots, x_n)$  be the demand (count) time series observed until time $n$. 
Similarly, let  $\mathbf{e}_n = (e_1, \dots, e_t)$ be the  respective residuals of a predictor dedicated to forecast $x_t$  ($t=1, \dots, n$).
In turn, let $X_{n+1}$ be the random variable that represents the next and unknown future value of the demand time series; 
and let $E_{n+1}$ be the respective error to be committed by the predictor. 
In fact, the challenge faced here is to provide a good forecast for the value of the time series at $(n+1)$,  $x_{n+1}$, in light of the observed histories $\mathbf{x}_n$ and $\mathbf{e}_n$. 

\subsection{Gamma--Poisson count model}
\label{sec:Gamma--Poisson count model}

In general terms, the Gamma--Poisson model arises when modeling count data for which the Poisson distribution serves as the likelihood and the Gamma distribution is chosen as the conjugate prior over the Poisson rate parameter $\lambda$ \cite{west2006bayesian, minois2017deal, turchetta2024time}.

Thus, consider $[X_{t+1} | \Lambda = \lambda_t] \sim \text{Poisson}(\lambda_t)$, in which $\lambda_t$ is an instance of the unknown rate of demand occurrences $\Lambda$ at $t$. 
In other terms, $\lambda_t$ governs the probabilistic mechanism underlying the next future and thus the uncertain value of $\mathbf{x}_t$, $X_{t+1}$.\prafC{use $X$ para a variável, $x$ para a respectiva observação, $\Lambda$  para a quantidade desconhecida da taxa e $\lambda$ para a respectiva instancia...ok} 
Assume the prior:

\[
\Lambda \sim \mathrm{Gamma}(\alpha_\Lambda, \beta_\Lambda),
\]
with the following general probability density function at time $t$:

\begin{equation}    
p(\lambda_t|\alpha_\Lambda, \beta_\Lambda) = \frac{\beta_\Lambda^{\alpha_\Lambda}}{\Gamma(\alpha_\Lambda)} \lambda_t^{\alpha_\Lambda-1} e^{-\beta_\Lambda \lambda_t}, \quad \lambda_t > 0.
\end{equation}
The pair $(\alpha_\Lambda, \beta_\Lambda)$ represents the uncertainty of the forecaster prior to the demand time series  $\mathbf{x}_t$. 
At time $t$ one has the following likelihood for an instance of $X_{t+1}$, say $x_{t+1}^*$:
\begin{equation}
    \label{lik_x(t+1)|l_t}
    p(x_{t+1}^* \mid \lambda_t) = \frac{e^{-\lambda_t} \lambda_t^{x_{t+1}^*}}{x_{t+1}^*!}.
\end{equation}

Using Bayes' theorem in light of $\mathbf{x}_t$, one has the posterior \cite{degroot2005optimal}:

\[
[\Lambda \mid \mathbf{x}_t] \sim \text{Gamma}\left(\alpha_\Lambda + \sum_{i=1}^t x_i,\ \beta_\Lambda + t\right),
\]\prafC{referencie DeGroot}
leading to the posterior mean forecast for $x_{t+1}^*$:
\begin{equation}        \label{eq:bayesian_forecast_x(t+1)}
\hat{\lambda}_{t+1} = \widehat{E(\Lambda \mid \mathbf{x}_t)}=\frac{a_\Lambda + \sum_{i=1}^t x_i}{b_\Lambda + t}
\end{equation}
\prafC{apresento a fórmula da média da Gamma, uso como símbolo para  a predição $\hat{\lambda}$. É dela que tiramos a previsão futura para $\Lambda$ e usamos como previsão, certo?? ...}

\begin{figure}[H]
    \centering
    \includegraphics[width=\linewidth]{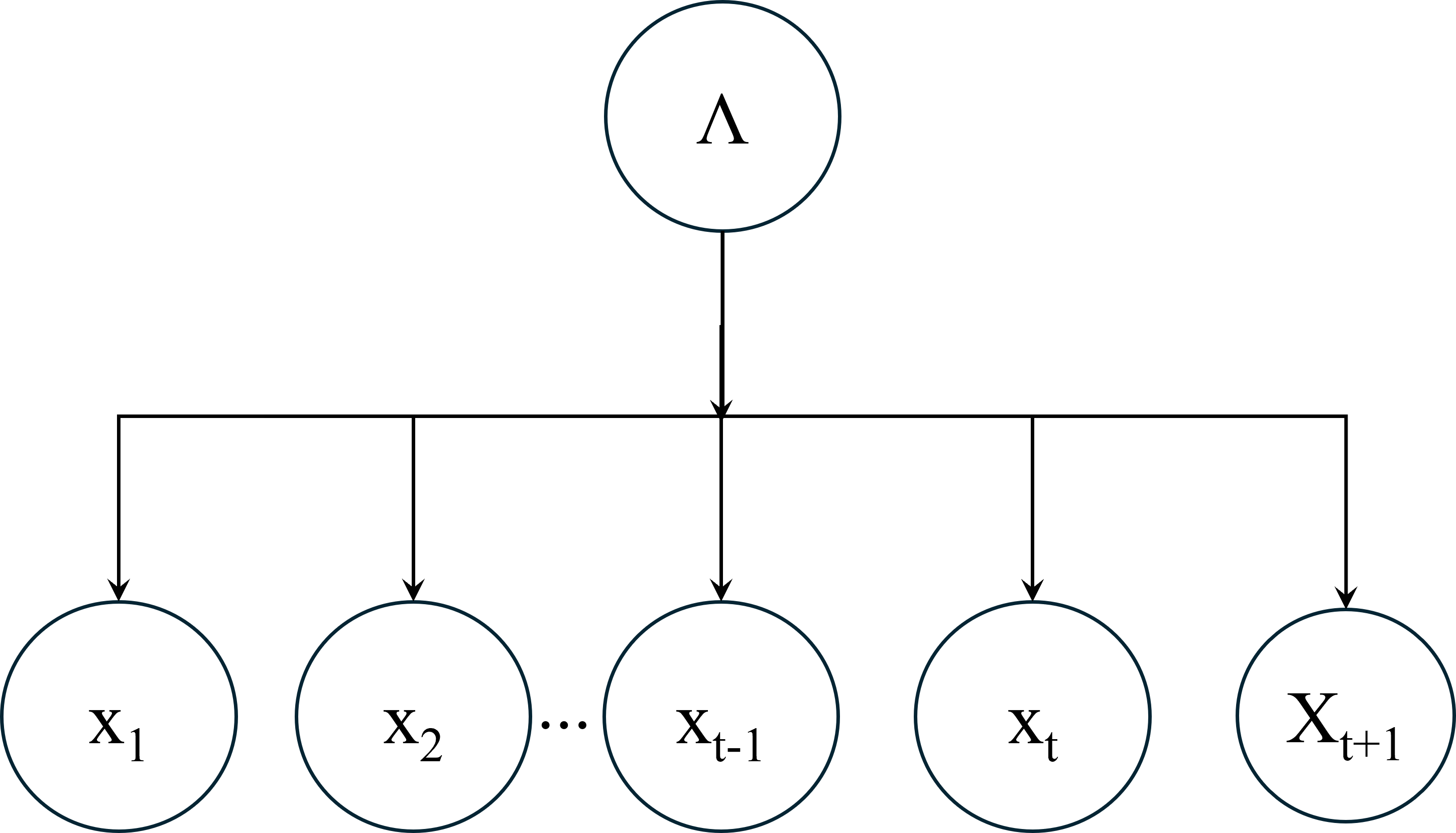}
\caption{Gamma-Poisson Bayesian net to model demand count processes. 
The likelihood function of the demand count process for time $(t+1)$ is $[X_{t+1}|\Lambda = \lambda_t] \sim Poisson(\lambda_t)$. Prior to any observation, one has \(\Lambda \sim \mathrm{Gamma}(\alpha_\Lambda, \beta_\Lambda)\). At time $t$, one has the evidence $\mathbf{x}_t=(x_1, \dots, x_t)$, leading to the posterior \( [ \Lambda \mid \mathbf{x}_t ] \sim \text{Gamma}\left(\alpha_\Lambda + \sum_{i=1}^t x_i,\ \beta_\Lambda + t\right)\).  It allows the prediction of $X_{t+1}$  with  $\hat{\lambda}_{t+1} = \widehat{E(\Lambda \mid \mathbf{x}_t)}=\frac{a_\Lambda + \sum_{i=1}^t x_i}{b_\Lambda + t}$.}\label{fig:Gamma_Poisson}
\end{figure}

This conjugacy greatly simplifies sequential inference and has been 
used in demand forecasting and clinical trial recruitment prediction \cite{dzupire2018poisson}. 
Figure \ref{fig:Gamma_Poisson} summarizes the demand count process with a Bayesian net graph. 
Thus, $[\Lambda|\mathbf{x}_t]$ encapsulates the long-term memory of the process: Eq \eqref{eq:bayesian_forecast_x(t+1)} provides an average between virtual and empirical evidence regarding the full count history $\mathbf{x}_t$ $(t=1, \dots, n)$. 
In turn, the autocorrelation underlying $\mathbf{x}_t$ is encapsulated by the distribution of $[\Lambda|\mathbf{x}_t]$. \prafC{referencie aqui a fórmula que pedi pra incluir a pouco ...}

\subsection{Gamma--Log-Normal error model}
\label{sec:Gamma--Log-Normal error model}

The Gamma-Log-Normal model may arise when modeling multiplicative residuals of a predictor (the ratio between the target values and their respective forecasts). 
Figure \ref{fig:bayesiannet}\prafC{não há imagem com o nome passado. ok} \prafC{a imagem da figura se dedica apenas à modelagem dos resíduos ...\par Creio que será melhor mesmo levar as redes Bayesianas de cada série (demandas e resíduos) para suas respectivas seções ... tem que atualizar a imagem ... Teremos $e_t$ e $E_{t+1}$ ... Nâo será $L$ mas sim $T$, nem $W$ mas sim $M$} shows the baseline network of the demand predictor error:
\begin{figure}[H]
    \centering    \includegraphics[width=0.9\linewidth]{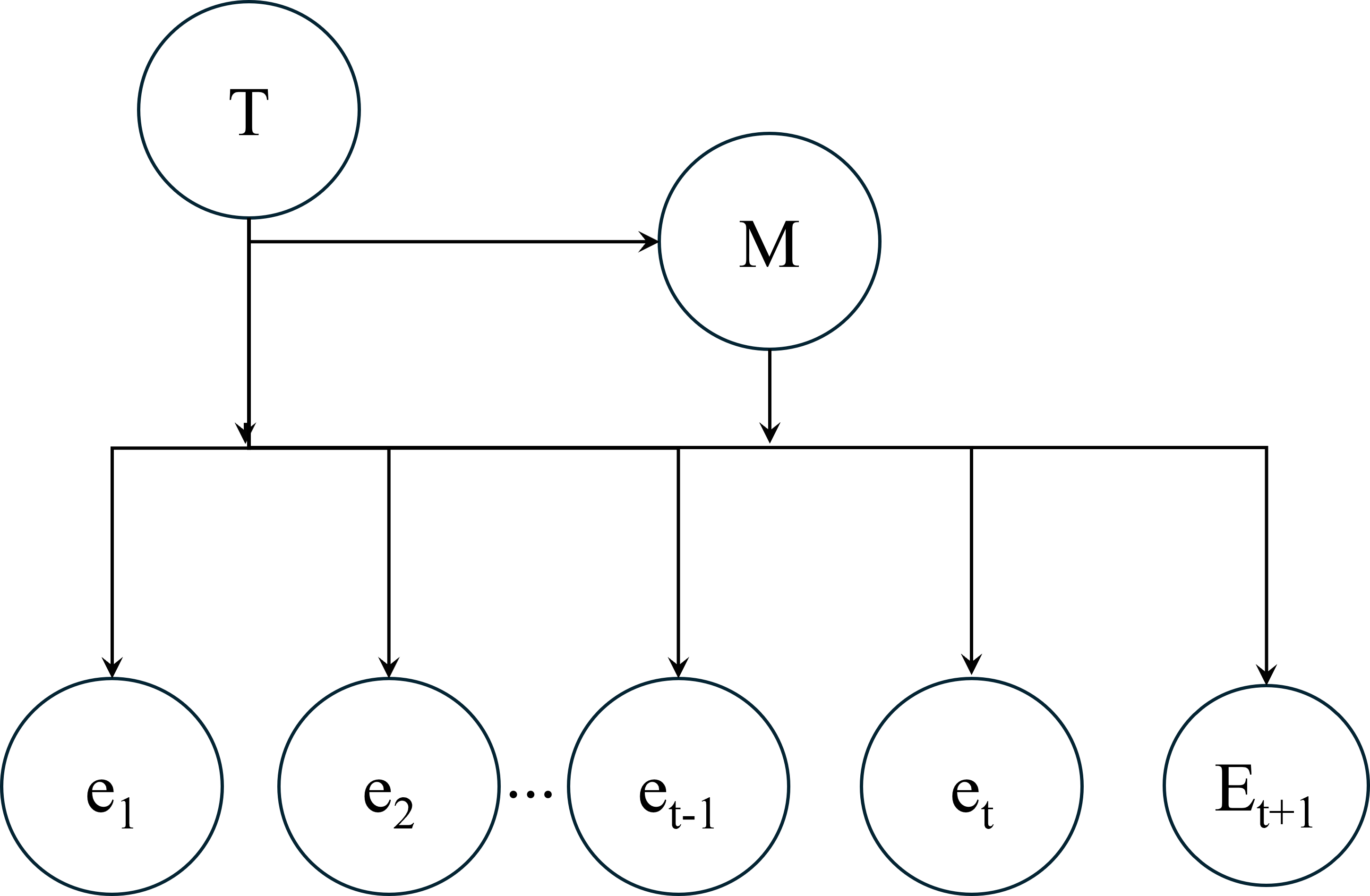}
\caption{Gamma–Log-Normal Bayesian net graph of the multiplicative error with respect to the demand predictor.
The likelihood function of the error series for time $(t+1)$ is $[E_{t+1}|M = \mu_t, T=\tau_t] \sim \mathcal{N}(\mu_t, \tau_t)$. 
Prior to any observation, one has \(T \sim \mathrm{Gamma}(\delta_T, \eta_T)\) and 
$[M =| T=\tau_t] \sim \mathcal{N}(m_M, \kappa_M\tau_t)$.
At time $t$, one has the evidence $\mathbf{e}_t=(e_1, \dots, e_t)$, leading to the 
prediction of $E_{t+1}$  with  $\hat{\mu}_{t+1} = \widehat{E(M \mid \mathbf{e}_t)}=\frac{\kappa_M m_M +  \sum_{i=1}^t e_i}{\kappa_M + t}$.}
    \label{fig:bayesiannet}
\end{figure}

It has been usual to suppose the logarithm of these residuals follows a Normal distribution \cite{Firmino2014}, \cite{Mosleh1986}.
Thus, the Normal distribution serves as the likelihood.
In this way, let 
\begin{equation}
\label{eq:log p1_additive_reridual}
e_t = \log_{p1} \left(\frac{x_t}{\hat{\lambda}_t}\right)= \log_{p1} (x_t) - \log_{p1} ( \hat{\lambda}_t)    
\end{equation}\prafC{talvez seja mesmo interessante usarmos o índice $t$ ao apresentarmos $X$, $x$ e $\lambda$...} be the logarithm-based additive residual of the forecast $\hat{\lambda}_t$ of $x_t$, such that $\log_{p1}(y) = \log(y+1)$. 
The $\log_{p1}(\cdot)$ function prevents the cases in which $x_t$ or $\hat{\lambda}_t$ vanish.
Further, let $E_{t+1}$ be the  random error of the count predictor for $x_{t+1}$, such that 
\begin{equation*}
[E_{t+1}|M = \mu_t, T = \tau_t] \sim \mathcal{N}(\mu_t, \tau_t).  
\end{equation*}
Here, $\mu_t$ and $\tau_t$ represent instances of the uncertain mean ($M$) and precision ($T$) of the predictor error at $t$, respectively \footnote{It is well known that the precision is the inverse of the variance \cite{degroot2005optimal} ($\tau_t = \frac{1}{\sigma_t^2}$ in which $\sigma_t^2$ is an instance of the uncertain variance of the predictor error at $t$).}. 
Therefore, similarly to $\lambda_t$ for $X_{t+1}$, the pair $(\mu_t, \tau_t)$ governs the probabilistic mechanism underlying the next future and thus uncertain value of $\mathbf{e}_t$, $E_{t+1}$.

Thus, the joint normal-Gamma distribution can be chosen as the conjugate prior over the Normal 
parameters. 
Specifically, one can use a Gamma prior for $T$:
\[
T \sim \text{Gamma}(\delta_T, \eta_T),
\]
and, conditionally on an instance of $T$, one can specify a Normal distribution for  $M$:
\[
[M \mid T=\tau_t] \sim \mathcal{N}\!\left(m_M, 
{\kappa_M \tau_t}\right).
\]

Hence, the joint prior distribution for $(M, T)$ follows a Normal--Gamma conjugate family, in turn conjugate with the normal likelihood of $E_{t+1}$.\praf{creio que seria muito bom trazermos as redes bayesianas dessa e da subseção anterior ... o leitor vai entender melhor}

\prafC{falta apresentar a distribuição de $\mu$ok}
The pair $(\delta_T, \eta_T)$ reflects the forecaster's uncertainty regarding the error precision $T$ prior to the predictor residuals time series \(\mathbf{e}_t=(e_1, \dots, e_t)\). 
At time $t$ one has the following likelihood for an instance of the error at $(t+1)$, $E_{t+1}$, say $e_{t+1}^*$:
\begin{equation} 
\label{eq:log_residual_likelihood}
p(e_{t+1}^* \mid \mu_t, \tau_t) \propto 
\exp\left(-\frac{\tau_t}{2}  (e_{t+1}^* - \mu_t)^2 \right).
\end{equation}

The 
Using Bayes' theorem in light of $\mathbf{e}_t$, one has the posteriors \cite{degroot2005optimal}:

\begin{equation}
[T \mid \mathbf{e}_t] \sim \text{Gamma}\left(\delta_T + \frac{t}{2},\ \eta_T + \frac{1}{2} \sum_{i=1}^t (e_i - \mu)^2 \right)
\end{equation}
\prafC{estou na dúvida sobre $\mu$...}
and 

\begin{equation}
\label{eq:M_given_tau_residuals}
[M \mid \tau_t, \mathbf{e}_t] \sim \mathcal{N}\!\left(
\frac{\kappa_M m_M + t \bar{e}_t}{\kappa_M + t},\ 
{(\kappa_M + t)\tau_t}
\right),
\end{equation}
in which $\bar{e}_t = \frac{1}{t}\sum_{i=1}^t {e}_i$. 
After dealing with the uncertainty underlying $T$ in Eq \eqref{eq:M_given_tau_residuals} (generically instantiated with $\tau_t$), one achieves the posterior mean forecast \cite{degroot2005optimal} for $e_{
t+1}^*$:
\prafC{falta explicitar a forma da posterior de $\mu$ ok}

\begin{equation}        \label{eq:bayesian_forecast_e(t+1)}
\hat{\mu}_{t+1} = \widehat{E(M \mid \mathbf{e}_t)} = \frac{\kappa_M m_M + t\bar{e}_t}{\kappa_M + t}.
\end{equation}

This model is generically useful in hierarchical settings and Bayesian regression models. 
It provides a way to encode uncertainty about the noise level in the data \cite{gelman2013bayesian}. 
$[M|\mathbf{e}_t]$ encapsulates the long-term memory of the residuals process: Eq \eqref{eq:bayesian_forecast_e(t+1)} provides a weighted average between virtual and empirical evidence regarding the full residuals history $\mathbf{e}_t$ $(t=1, \dots, n)$.

\section{Proposed Method}

The proposed framework combines the two conjugate models previously presented. 
It involves a residual correction mechanism to improve demand forecasts, taking healthcare applications as examples. 
The method unfolds in three key stages: 
(i) a Gamma-Poisson base model for demand estimation (Section \ref{sec:Gamma--Poisson count model}), 
(ii) a log-based transformation of multiplicative residuals of the base model to approximate a Gaussian distribution, and a respective Gamma-Normal residual model forecast (Section \ref{sec:Gamma--Log-Normal error model}); and  
(iii) a boosting model that sums the forecasts from both the count and error models. 

Thus $\mathbf{x}_t$ \prafC{na seção anterir vc usou x, não u. ok} is the observed oncological demand series until time \( t \). 
In light of $\mathbf{x}_t$ we forecast $x_{t+1}$ with Eq \eqref{eq:bayesian_forecast_x(t+1)}. 
Further, to improve model accuracy, in light of the base-model previous residuals, $\mathbf{e}_t$, we predict the next one, $e_{t+1}$, via Eq \eqref{eq:bayesian_forecast_e(t+1)}.
Finally, to boost the  forecast of $x_{t+1}$, we consider: 

\begin{equation}
\label{eq: x_hat + x_hat}
\hat{x}_{t+1} = 
\hat{\lambda}_{t+1} + \max[0, \quad \exp(\hat{\mu}_{t+1})-1].
\end{equation}
\prafC{\textcolor{red}{acho que não precisamos desse $^{corrected}$}..\par ademir: Eu mantive para destinguir da previsao pela média e a pervisão pela média com o valor corrigido no boosting\par praf: agora ficou redondo...}
It is worthwhile to remember that $e_t$, and thus $\hat{\mu}_{t+1}$, are based on $\log_{p1}$ scale (see Eq \eqref{eq:log p1_additive_reridual}), leading to the inverse transform ($\exp(\hat{\mu}_{t+1})-1$). 
In turn, the presence of $\max(\cdot)$ function in Eq \eqref{eq: x_hat + x_hat} prevents float point numerical problems. 
The correction of the proposed model aims to mitigate systematic biases in the forecasts by accounting for persistent discrepancies between model predictions and actual observations while allowing online updating. 
This is the pseudo code of the proposed method:\prafC{atualizar os símbolos do algoritmo de acordo com os da seção anterior ...\par o algoritmo deve envolver as equações \eqref{eq:bayesian_forecast_x(t+1)} e \eqref{eq:bayesian_forecast_e(t+1)}} \praf{ como está parece errado (rever se está tudo ok no código): a média dos residuos no log deve ser calculada primeiro ... só depois que se calcula a inversa... (permutar linhas 10 e 12) a correção da linha 13 é aditiva (envolve soma, não subtração). a eq da linha 8 tá inconpleta ...}

\begin{algorithm}[H]
\caption{Sequential Bayesian Boosting Forecast (Gamma-Poisson + Gamma-Log-Normal)}
\begin{algorithmic}[1]
\REQUIRE Training data $x_{1:T_{\text{train}}}$, Test data $x_{T_{\text{train}}+1:T_{total}}$, Hyperparameters $\kappa_M, m_M$
\STATE Initialize Gamma-Poisson prior using first observation $x_1$:
    \IF{$x_1 > 0$}
        \STATE $\beta_{\Lambda} \gets 10^{-\lfloor \log_{10}(x_1) \rfloor}$, $\alpha_{\Lambda} \gets x_1 \cdot \beta_{\Lambda}$
    \ELSE
        \STATE $\beta_{\Lambda} \gets 0.0001$, $\alpha_{\Lambda} \gets \beta_{\Lambda} \cdot 100$
    \ENDIF
\STATE Initialize Error Model statistics: $S_e \gets 0$ (Sum of residuals), $N \gets 0$ (Count)

\FOR{$t = T_{\text{train}}+1$ to $T_{total}$}
    \STATE \textbf{Step 1: Base Model Forecast}
    \STATE $\hat{\lambda}_t \gets \alpha_\Lambda / \beta_\Lambda$

    \STATE \textbf{Step 2: Error Model Forecast (Eq. \ref{eq:bayesian_forecast_e(t+1)})}
    \STATE $\hat{\mu}_t \gets \frac{\kappa_M m_M + S_e}{\kappa_M + N}$

    \STATE \textbf{Step 3: Boosted Forecast (Eq. \ref{eq: x_hat + x_hat})}
    \STATE $\hat{x}_t \gets \hat{\lambda}_t + \max\left(0, \exp(\hat{\mu}_t) - 1\right)$

    \STATE \textbf{Step 4: Observation and Update}
    \STATE Observe actual value $x_t$
    \STATE Compute residual: $e_t \gets \log_{p1}(x_t) - \log_{p1}(\hat{\lambda}_t)$ \COMMENT{Eq. \ref{eq:log p1_additive_reridual}}
    \STATE Update Error statistics: $S_e \gets S_e + e_t$, $N \gets N + 1$
    \STATE Update Base prior: $\alpha_\Lambda \gets \alpha_\Lambda + x_t$, $\beta_\Lambda \gets \beta_\Lambda + 1$
\ENDFOR
\end{algorithmic}
\end{algorithm}

The pseudocode outlines the proposed sequential Bayesian forecasting method with residual correction. 
The algorithm begins by initializing the Gamma prior parameters $\alpha$ and $\beta$ based on the first observation. 
At each time step, a prediction $\hat{\lambda}_t$ is generated from the current posterior mean. 
The residual is calculated on a log-transformed scale to stabilize variance and then back-transformed to the original scale. 
An accumulated average of the residuals up to time $t$ is used to correct systematic bias in the forecast. 
Additionally, a reset mechanism is triggered if the residuals indicate a significant and abrupt change in the underlying data pattern, ensuring adaptability. 
After each observation, the posterior parameters are updated by incorporating the new data point, maintaining a continuous Bayesian learning process throughout the forecasting horizon, which characterizes an online learning paradigm.
\prafC{destacar o paradigma de aprendizado online ...ok}

\section{Experiments}
\label{sec:methodology}

This section presents the experimental setup used to evaluate the proposed Bayesian forecasting model with residual-based boosting. Here, we describe the datasets and introduce the baseline models.

\subsection{Datasets}
\label{subsec:datasets}
The time series used as case studies were retrieved from a weak oncological demand in the São Vicente de Paula Hospital, located in Cariri, Ceará, Brazil. 
This institution provided the top 10 series, i.e., the most required oncology services, divided by CID (mnemonic for International Statistical Classification of Diseases and Related Health Problems). 
The data were collected between January 8th, 2023, and December 1st, 2024. 
Table~\ref{tab:dataset_summary} summarizes the CIDs dataset, including the specification of the disease, weak amount of demands, maximum, minimum, mean value, and standard deviation. 
One can see Prostate cancer is the most prevalent and widespread, followed by Breast and Rectal canser.

\begin{table}[H]
\caption{Descriptive statistics of the oncology demand time series (ordered by mean) taken as case studies, obtained from São Vicente de Paula Hospital, located at Cariri, Ceará, 
Brazil, between 2023-01-08 and 2024-12-01.}
\label{tab:dataset_summary}
\resizebox{\textwidth}{!}{  
\begin{tabular}{llrrrrr}
\toprule
\multirow{2}{*}{CID} & \multirow{2}{*}{Description} & \multirow{2}{*}{Size} & \multicolumn{4}{c}{Observed Numerical Values}\\ \cline{4-7}
    &             &      & Max & Min & Mean & Std Dev \\
\midrule
C61 & Prostate cancer & 100 & 258 & 43 & 153.00 & 49.62 \\
C509 & Breast cancer & 100 & 196 & 46 & 104.68 & 34.83 \\
C20 & Rectal cancer & 100 & 42 & 3 & 19.42 & 8.09 \\
C109 & Malignant neoplasm of the oropharynx & 100 & 51 & 1 & 17.11 & 11.58 \\
C539 & Malignant neoplasm of the cervix & 100 & 39 & 0 & 15.01 & 7.81 \\
C508 & Breast cancer with invasive lesion & 97 & 48 & 0 & 14.44 & 12.08 \\
C349 & Bronchus and lung cancer & 100 & 33 & 1 & 14.17 & 7.05 \\
C159 & Malignant neoplasm of the esophagus & 99 & 35 & 1 & 13.41 & 8.27 \\
C189 & Malignant neoplasm of the colon & 100 & 23 & 2 & 11.43 & 4.99 \\
C169 & Malignant neoplasm of the stomach & 100 & 22 & 1 & 10.96 & 4.51 \\
\bottomrule
\end{tabular}
}
\end{table}

\prafC{the table is not ordered by mean ...ok}
\subsection{Baselines}
We compare the proposed model against five baseline approaches commonly used in time series forecasting: Naïve forecasting, Autoregressive Linear Regression, Extreme Gradient Boosting (XGBoost),  ARIMA, and LSTM neural network. 
The Naïve forecasting method assumes that the best prediction for the next period is simply the last observed value. Despite its simplicity, it often serves as a surprisingly competitive baseline, especially when the underlying data-generating process resembles a Naïve forecasting \cite{hyndman2018forecasting}. 
The autoregressive linear regression model is a simple yet effective approach for time series forecasting. 
By modeling the current value of the target variable as a linear combination of its past observations, specifically using the significant lags obtained by partial autocorrelation tests \cite{hyndman2018forecasting}. 
The significant lags were selected at a 5\% significance level to reduce overfitting and emphasize relevant temporal dependencies. 
The linear regression model was implemented using the Python library \texttt{scikit-learn} \cite{pedregosa2011scikit}, a widely adopted open-source machine learning toolkit known for its ease of use and comprehensive suite of algorithms.

The Extreme Gradient Boosting (XGBoost)\prafC{vc não mencionou XGBoost no parágrafo anterior... ok \textcolor{red}{ nos impede de usar outros modelos, como SVR, MLP e LSTM? Suspeito que ganhamos de redes neurais sem maiores problemas} ... na verdade nada nos impede, inclusive esse experimento foi feito com LSTM tbm, mas o resultado foi muito ruim e decidimos tirar do estudo} model is a machine learning technique based on gradient boosting decision trees. 
It is designed to handle complex, nonlinear relationships and is known for its efficiency, scalability, and predictive accuracy in a wide range of regression tasks \cite{chen2016xgboost}. 
In this study, the implementation from the Python library \texttt{xgboost}\cite{xgboost_python}\prafC{referencie} is used, with an autoregressive input structure based on the same significant lags selected by partial autocorrelation analysis. 
Hyperparameters were tuned through a grid search procedure using time-series cross-validation (3 folds), optimizing the mean absolute error (MAE) on the training data to enhance generalization and prevent overfitting. 
The main parameters  include: \texttt{n\_estimators} (200–500), \texttt{max\_depth} (2–4), \texttt{learning\_rate} (0.03–0.05), \texttt{subsample} and \texttt{colsample\_bytree} (0.8–1.0), and \texttt{gamma} (0–1).

The Long Short-Term Memory (LSTM) net is a recurrent neural network architecture specifically designed to capture long and short range temporal dependencies in sequential data \cite{hochreiter1997long}.
In this study, the input was structured using sliding windows of past observations to preserve temporal ordering, and the network was trained to iteratively predict future values.
Hyperparameters were optimized through a grid search procedure with time-series cross-validation, ensuring better generalization and reducing the risk of overfitting.
The final model employed an autoregressive structure based on significant lags identified via the Partial Autocorrelation Function (PACF). In this recursive framework, each new prediction was fed back as input to generate subsequent values, ensuring the approach is directly comparable with other autoregressive models.\prafC{a ordem auto-regressiva veio do PACF? informe.}

Lastly, the Autoregressive Integrated Moving Average (ARIMA) model serves as a classical statistical method that captures linear dependencies and temporal structures in time series. 
It combines autoregressive (AR) components, which model the dependence on past values, with moving average (MA) components, which capture the dependence on past forecast errors \cite{box2015time}. 
In this study, model parameters are automatically selected based on the training data using the Python library \textit{pmdarima}, which implements the Box–Jenkins methodology for time series modeling and forecasting \cite{smith2017pmdarima}. 
The model order is determined via a stepwise search that minimizes the Akaike Information Criterion (AIC).

\subsection{Evaluation Measures}
The performance of the models was evaluated using the Percentage of Correct Direction (POCID).This measure quantifies the proportion of instances in which the predicted trend matches the actual trend, rewarding correct directional predictions and penalizing incorrect ones \cite{asadi2012hybrid}. 
This is particularly suitable for applications where capturing the direction of movement plays an important role. 
Formally, POCID is defined as:
\begin{equation}
   \text{POCID} = \frac{1}{n} \sum_{t=2}^n \mathbb{I} \big[ \text{sign}(\hat{x}_t - x_{t-1}) = \text{sign}(x_t - x_{t-1}) \big],
\end{equation}
in which $\mathbb{I}(\cdot)$ denotes the indicator function, returning 1 when the predicted and actual changes share the same sign (i.e., trend direction) and 0 otherwise.
\section{Results and Discussion}



\begin{figure}[H]
    \centering
    \includegraphics[width=0.9\textwidth]{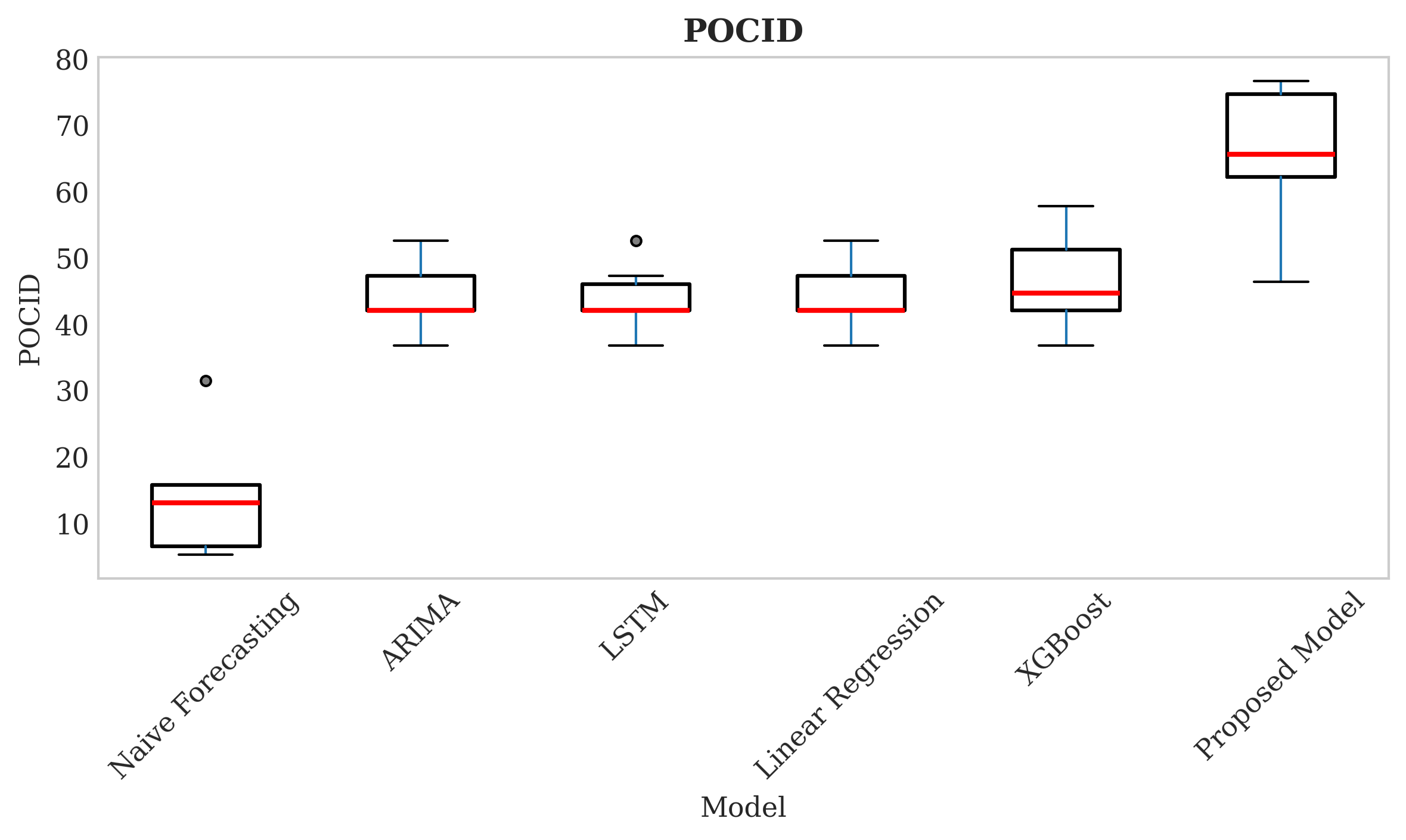}
    \caption{Boxplot of POCID across models: ARIMA, Linear Regression, Naïve forecasting, XGBoost, LSTM, and Proposed Model
    with respect to the oncology demand data taken as case studies, obtained from São Vicente de Paula Hospital, located at Cariri, Ceará, 
Brazil, between 2023-01-08 and 2024-12-01}
    \label{fig:boxplot_pocid}
\end{figure}

Figure~\ref{fig:boxplot_pocid} presents the distribution of POCID values for all evaluated models per formalism.
Among the evaluated models, the proposed model achieved the highest POCID values across most oncology demand series, indicating a superior capacity to correctly predict the direction of change in demand trends. The Naïve forecasting model, despite its simplicity, showed the worst performance in the POCID term, indicating that relying solely on recent past values is insufficient to capture directional changes, particularly in volatile patterns. ARIMA,  
 LSTM and Linear Regression models tended to present lower POCID scores, revealing difficulties in adapting to the structural variability and stochastic components observed in the series. 
It is expected that the LSTM performance is proportional to the sample size, justifying the modest results of these recurrent neural nets in the current study cases.
XGBoost exhibited a mixed performance: although it did not consistently outperform the Proposed Model, it reached high POCID values in specific cases, reflecting its capacity to capture certain nonlinear dependencies. 

As shown in Figure~\ref{fig:boxplot_pocid}, the Proposed Model not only achieves the highest median POCID but also displays low dispersion\prafC{não vejo isso do grafico. O nosso parece ser o mais disperso}, indicating both accuracy and stability across different time series. 
Naïve forecasting presents a slightly lower median and high variability\prafC{nossa variabilidade paerec ser maior}, while ARIMA, Linear Regression, and XGBoost exhibit wider interquartile ranges\prafC{nao vejo isso} and lower central tendencies, underscoring their reduced consistency. 
These results demonstrate that the integration of prior information with residual-based boosting, as implemented in the Proposed Model, provides an advantage in correctly identifying demand trend directions in nonstationary healthcare time series.

In terms of scale-sensitive metrics, the proposed model markedly reduced errors against the parametric and tree-based baselines. Across the ten series, it achieved median MAPE reductions of 74.5\% vs ARIMA, 77.6\% vs Linear Regression, 75.7\% vs LSTM, and 74.7\% vs XGBoost, alongside median Theil’s U reductions of 73.9\% vs ARIMA, 83.0\% vs Linear Regression, 83.1\% vs LSTM, and 71.6\% vs XGBoost. \prafC{e sobre LSTM? se LSTM foi tão ruim, usamos o tamanho da amostra como justificativa ...}
The largest gains were observed against Linear Regression in C109 (MAPE --85.8\%; U --97.0\%) and against XGBoost in C349 (MAPE --79.9\%) and C509 (U --97.1\%). \prafC{não me recordo de termos dito se esses modelos alternativos operaram sob regime offline ou online ... certifique-se de ter dito ... traga a discussão sobre isso pra cá tb...}
Compared with the naïve Random Walk (U=1), the corrected model sometimes delivered substantial efficiency gains (e.g., C169: U=0.39, --61\%), but also exhibited cases with higher U (notably C109 and the outlier C508). 
Overall, these results indicate that the correction substantially improves accuracy and efficiency over ARIMA, Linear Regression, and XGBoost, whereas its advantage over the naïve benchmark remains series-dependent.\prafC{ADEMIR: Professor Paulo, adicionei esse parágrafo para entregar uma análise com relação a MAPE e Theil dos resultados.\par praf: show!}

\prafC{\textcolor{red}{observando os resultados, pensei que talvez os revisores sintam falta de alguma medida que destaque o desempenho além da tendência ... qual foi a segunda medida em que nos saimos melhor?}Anteriormente tinhamos 4 metricas, mas apenas o POCID teve resultado estatisticamente significativo}

\subsection{Statistical Comparison}

To infer which differences in POCID between models are statistically significant, we applied the Diebold-Mariano (DM) test \cite{diebold1995}. 
For each time series $s$ and each pair of models, say $(m_a, m_b)$, the DM statistic is computed as:

\begin{equation}
DM(m_a, m_b) = \frac{\bar{d}}{\sqrt{\hat{\sigma}_d^2 / n}},
\end{equation}
in which 
\begin{equation}
\bar{d} = \frac{1}{n}\sum_{t=1}^n d_t, \quad d_t = L(e_{a,t}) - L(e_{b,t}).
\end{equation}
\prafC{apresente a equação}
It has been considered here the squared root loss component $L(e_{a,t}) = (x_t - \hat{x}_t^y)^2,$
in which $\hat{x}_t^y$ represents the forecast of the model $m_y$ for $x_t$.  
In turn,  $\hat{\sigma}_d^2$ is the  variance of the loss differences sample $(d_1, \dots, d_n)$.  
A negative and statistically significant $DM(m_a, m_b)$ indicates that model $m_a$ outperforms model $m_b$ for the given series in terms of MSE.

In this study, for each series, we selected the model that obtained the highest POCID and was statistically different\prafC{sugiro aqui que faça teste bicaudal. Se H0 for rejeitada, o sinal da estatística de DM sugere quem venceu. Sempre considere o nosso modelo como sendo o $m_a$ ... } according to the MSE-based DM test\prafC{só conferindo uma última vez, o DM test foi baseado no MSE, né?}.  
When no statistically significant difference was found, we applied a fallback criterion that prioritizes the \textit{Proposed Model}, except in cases of exact ties, which were labeled as \textit{Tie}.
Table \ref{tab:dm_results} \prafC{esses 0.0 pp da tabela estão confusos. Houve empate numérico entre o nosso e o segundo melhor? reveja isso. informe o sinal da estatistica de DM. Já ajustei o texto pra que H1 seja diferente. Assim, tu estarás fazendo um teste bicaudal... como sempre teremos o modelo proposto como sendo $m_a$, desejamos que DM difference, baseado no MSE, seja negativo ...} summarizes the best-performing model per series, the DM-based difference in MSE (in percentage points), and the decision source (statistically significant or fallback). 
One can see that the proposed method statistically beat the second best model in terms of MSE in 7 out of 10 cases, a remarkable result.

\begin{table}[H]
\centering
\caption{Best model per series according to the Diebold-Mariano test based on MSE (two-tailed test), where the Proposed Model is \textbf{$m_a$}. A negative value ($-$) for the DM statistic indicates that the Proposed Model outperformed the baseline (\textbf{$MSE_a < MSE_b$}).}
\label{tab:dm_results}
\begin{tabular}{ccc}
\toprule
\textbf{Series} & \textbf{Statistical Result} & \textbf{DM Stat. ($m_a$ vs $m_b$)} \\
\midrule
C109 & Proposed Wins & -10.53 \\
C159 & Inconclusive & -- \\
C169 & Inconclusive & -- \\
C189 & Proposed Wins & -7.44 \\
C20  & Proposed Wins & -1.70 \\
C349 & Proposed Wins & -2.07 \\
C508 & Inconclusive & -- \\
C509 & Proposed Wins & -5.26 \\
C539 & Proposed Wins & -14.30 \\
C61  & Inconclusive & -- \\
\bottomrule
\end{tabular}
\end{table}

\section*{Conclusion}

The results of this study highlight the attractiveness of incorporating Bayesian inference with correction mechanisms in time series forecasting exercises of cancer demand forecasting. 
Compared to alternative models such as XGBoost, ARIMA, Linear Regression, LSTN, and even the baseline Naïve forecasting, the boosting Bayesian model consistently achieved superior performance across multiple evaluation metrics. 
This reinforces the value of probabilistic reasoning and the integration of prior knowledge in addressing the uncertainty and variability inherent in real-world data.

Despite these promising results, several challenges were encountered throughout the modeling process. 
For instance, one challenge involved the model's parsimony, especially when working with relatively small datasets for each series. 
Ensuring fair comparisons among models with different assumptions and structures also required careful metric selection and normalization procedures. 
Furthermore, the sensitivity of certain evaluation metrics to outliers influenced the final assessments, demanding robust aggregation strategies.

Another important limitation lies in the sensitivity of the proposed model to the initialization of prior distributions. 
The selection of informative or non-informative priors can considerably affect the convergence and quality of posterior estimates, particularly when data is scarce or highly variable. 
Future studies might further explore systematic approaches for prior elicitation, potentially incorporating empirical Bayes techniques or domain expert input.


Future work will also explore the integration of short- and long-term memory mechanisms into the Bayesian inference framework, drawing inspiration from LSTM networks. 
By embedding memory units capable of capturing temporal dependencies and dynamic patterns over time, we aim to enhance the model's capacity to adapt to evolving data structures. 
This hybrid approach is expected to improve both the predictive power and the interpretability of Bayesian networks, especially in non-stationary or highly volatile time series environments.



\bibliographystyle{plain}
\bibliography{sbc-template}

\end{document}